\documentclass[conference]{IEEEtran}
\IEEEoverridecommandlockouts
\usepackage{cite}
\usepackage{amsmath,amssymb,amsfonts}
\usepackage{algorithmic}
\usepackage{graphicx}
\usepackage{textcomp}
\usepackage{xcolor}

\usepackage[utf8]{inputenc}
\usepackage{booktabs}
\usepackage{threeparttable}
\usepackage[hidelinks]{hyperref}

\def\BibTeX{{\rm B\kern-.05em{\sc i\kern-.025em b}\kern-.08em
    T\kern-.1667em\lower.7ex\hbox{E}\kern-.125emX}}
    
\DeclareRobustCommand*{\IEEEauthorrefmark}[1]{%
    \raisebox{0pt}[0pt][0pt]{\textsuperscript{\footnotesize\ensuremath{#1}}}}

\begin{document}

\title{PlantBiMoE: A Bidirectional Foundation Model with SparseMoE for Plant Genomes\\

}

\author{
    \IEEEauthorblockN{Kepeng Lin\textsuperscript{1,\dag}, Qizhe Zhang\IEEEauthorrefmark{2,3,\dag}, Rui Wang\IEEEauthorrefmark{2,3}, Xuehai Hu\IEEEauthorrefmark{2,3,*} and  Wei Xu\IEEEauthorrefmark{1,*} }
    \IEEEauthorblockA{\IEEEauthorrefmark{1} School of Electronic Information and Communications, Huazhong University of Science and Technology, Wuhan, 430074, China}
    \IEEEauthorblockA{\IEEEauthorrefmark{2} Hubei Hongshan Laboratory, Wuhan 430070, China}
    \IEEEauthorblockA{\IEEEauthorrefmark{3} College of Informatics, Agricultural Bioinformatics Key Laboratory of Hubei Province, \\
    Huazhong Agricultural University, Wuhan, 430070, China }
    \IEEEauthorblockA{\IEEEauthorrefmark{\dag} These authors contributed equally to this work }
    \IEEEauthorblockA{\IEEEauthorrefmark{*} Correspondence to: Xuehai Hu, Email to: huxuehai@mail.hzau.edu.cn, Wei Xu, Email to: xuwei@hust.edu.cn}
}

\maketitle

\begin{abstract}
Understanding the underlying linguistic rules of plant genomes remains a fundamental challenge in computational biology. Recent advances including AgroNT and PDLLMs have made notable progress although, they suffer from excessive parameter size and limited ability to model the bidirectional nature of DNA strands respectively. To address these limitations, we propose PlantBiMoE, a lightweight and expressive plant genome language model that integrates bidirectional Mamba and a Sparse Mixture-of-Experts (SparseMoE) framework. The bidirectional Mamba enables the model to effectively capture structural dependencies across both the forward and reverse DNA strands, while SparseMoE significantly reduces the number of active parameters, improving computational efficiency without sacrificing modeling capacity. We evaluated and tested our model on the Modified Plants Genome Benchmark (MPGB), an enhanced genomic benchmark, which consolidates 31 datasets across 11 representative tasks, with input sequence lengths ranging from 50 to 6,000 bp. Experimental results demonstrate that PlantBiMoE achieves the best performance on 20 out of 31 datasets and the average best when comparing with existing models. In summary, all above results demonstrate that our model can effectively represent plant genomic sequences, serving as a robust computational tool for diverse genomic tasks, while making substantive contributions to plant genomics, gene editing, and synthetic biology. The code is available at: https://github.com/HUST-Keep-Lin/PlantBiMoE
\end{abstract}


\begin{figure}[!b]
\centering
\includegraphics[width=0.50\textwidth]{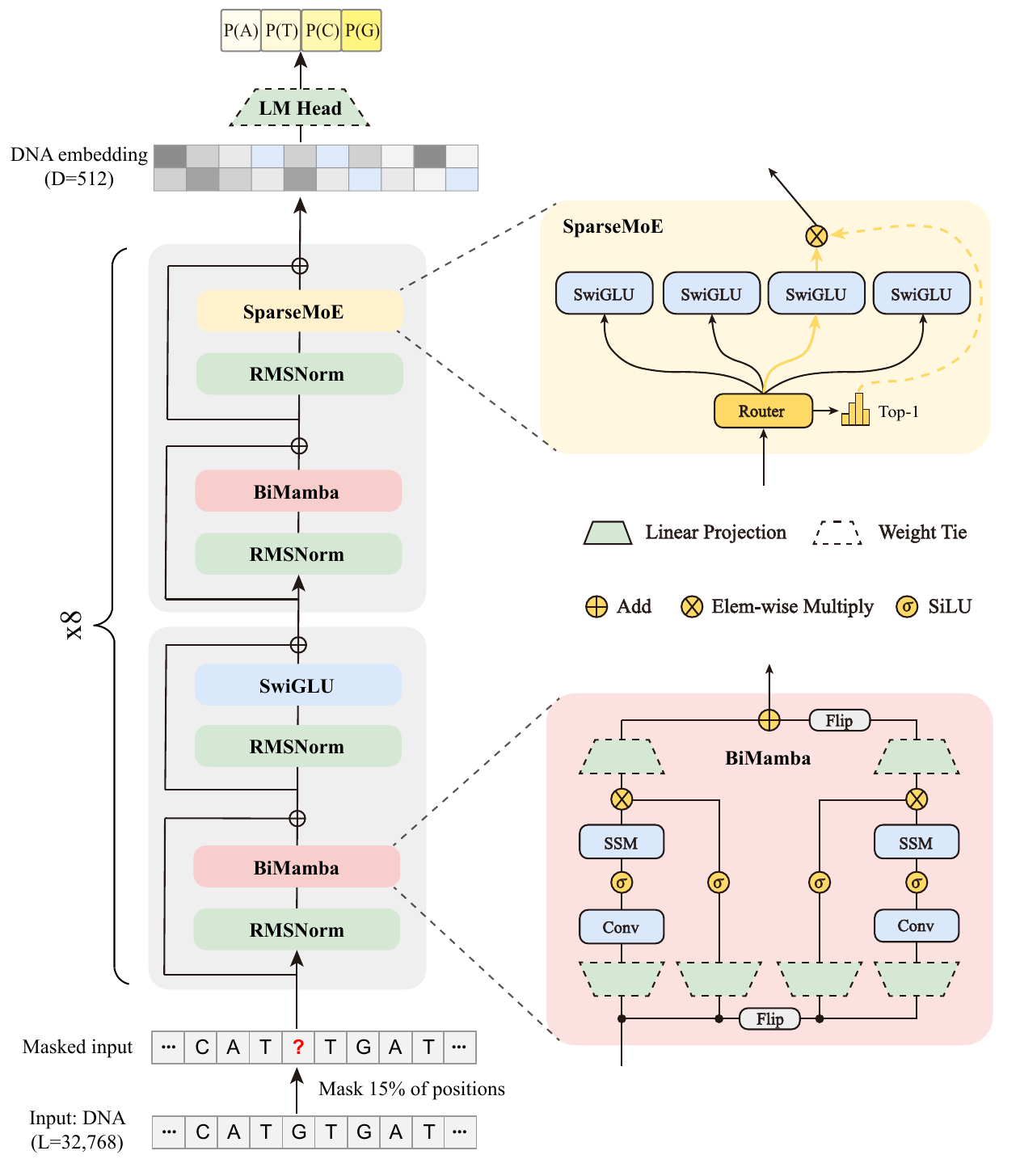}
\caption{Model architecture of PlantBiMoE.} 
\label{fig:f1}
\end{figure}

\section{Introduction}
The plant genome harbors a large number of sequence elements with both regulatory and structural functions. These elements are distributed not only coding sequence, but also non-coding regions, contributing to a wide range of gene regulatory processes, including transcription initiation, alternative splicing, polyadenylation signal recognition, mRNA stabilization, chromatin accessibility, and epigenetic modification\cite{CREs}\cite{nonEncoding}. Such regions typically exhibit uneven base composition, long-range dependencies, and complex cross-segment interactions, which pose significant challenges for traditional window-based models that lack global context modeling capabilities\cite{deepplantcre}. With the growing availability of high-throughput sequencing data across diverse plant species, there is an urgent need for a unified sequence modeling framework that supports cross-species generalization and multitask learning. Such a model would provide a semantic foundation for genome annotation, regulatory mechanism inference, and a variety of sequence-based genomic analyses.

In recent years, DNA language modeling has emerged as a powerful approach in computational genomics, particularly in the domains of animals and microorganisms. Pre-trained models such as DNABERT\cite{DNABERT}, DNABERT-2\cite{DNABERT2}, Nucleotide Transformer\cite{NT}, and Evo\cite{evo} have demonstrated remarkable cross-species generalization on tasks such as functional element recognition, sequence classification, and variant effect prediction. In contrast, the progress in plant genome language modeling lags significantly behind. Existing models are few in number and often focused on narrow species scopes or specific tasks, lacking generalizability and comprehensive coverage. AgroNT\cite{AgroNT} is the first large-scale pre-trained Transformer model for plant genomes, achieving strong performance in promoter strength prediction, splice site detection, and non-coding RNA classification. However, its parameter scale exceeds one billion, making it resource-intensive and difficult to deploy in typical research settings. Additionally, it inherits the standard BERT\cite{bert} architecture without incorporating DNA-specific structural priors such as strand symmetry or transcript boundary information. Subsequent PDLLMs\cite{PDLLMs} introduce lightweight variants (e.g., PlantDNAMamba, PlantDNAGPT) with reduced inference cost and initial support for multitask learning. Unfortunately, most adopt unidirectional modeling and heavily standardized feedforward blocks, which limits their ability to represent functionally dense and asymmetric regions. PlantCaduceus\cite{PlantCaduceus} further combines Mamba-based state space modeling with reverse-complement symmetry. While offering improved bidirectional representation, its modeling scope is confined to short-range windows (512 bp), making it insufficient for capturing long-range \textit{cis}-regulatory interactions.

To address these limitations, we here propose PlantBiMoE, a novel language modeling framework for plant genomes that integrates a bidirectional state space model\cite{Caduceus} with a Sparse Mixture-of-Experts SparseMoE architecture\cite{SparseMoE}. The bidirectional Mamba core enables efficient encoding of upstream and downstream strand information, while SparseMoE reduces parameter overhead and enhances adaptability to heterogeneous sequence regions. PlantBiMoE is pre-trained on whole-genome sequences from 42 representative plant species, with a total of 25.40B nucleic acid pairs. The final model contains 116M parameters. To rigorously evaluate PlantBiMoE, we introduce the Modified Plants Genome Benchmark (MPGB)—a modified benchmark derived from the Plants Genome Benchmark\cite{AgroNT}, comprising 31 datasets across 11 genomic tasks with input lengths ranging from 50 to 6,000 bp. Experimental results show that it consistently outperforms AgroNT and PDLLMs across most tasks, particularly excelling in cross-species prediction, long-context modeling, and epigenetic feature recognition. These results establish PlantBiMoE as a new paradigm for plant genomic sequence modeling and a versatile foundation for genome annotation and regulatory inference.

\section{MATERIALS AND METHODS}

\subsection{Dataset}

1) \textit{A pre-training dataset comprising genomic sequences from 42 plant species}

\begin{figure}[!b]
\centering
\includegraphics[width=0.50\textwidth]{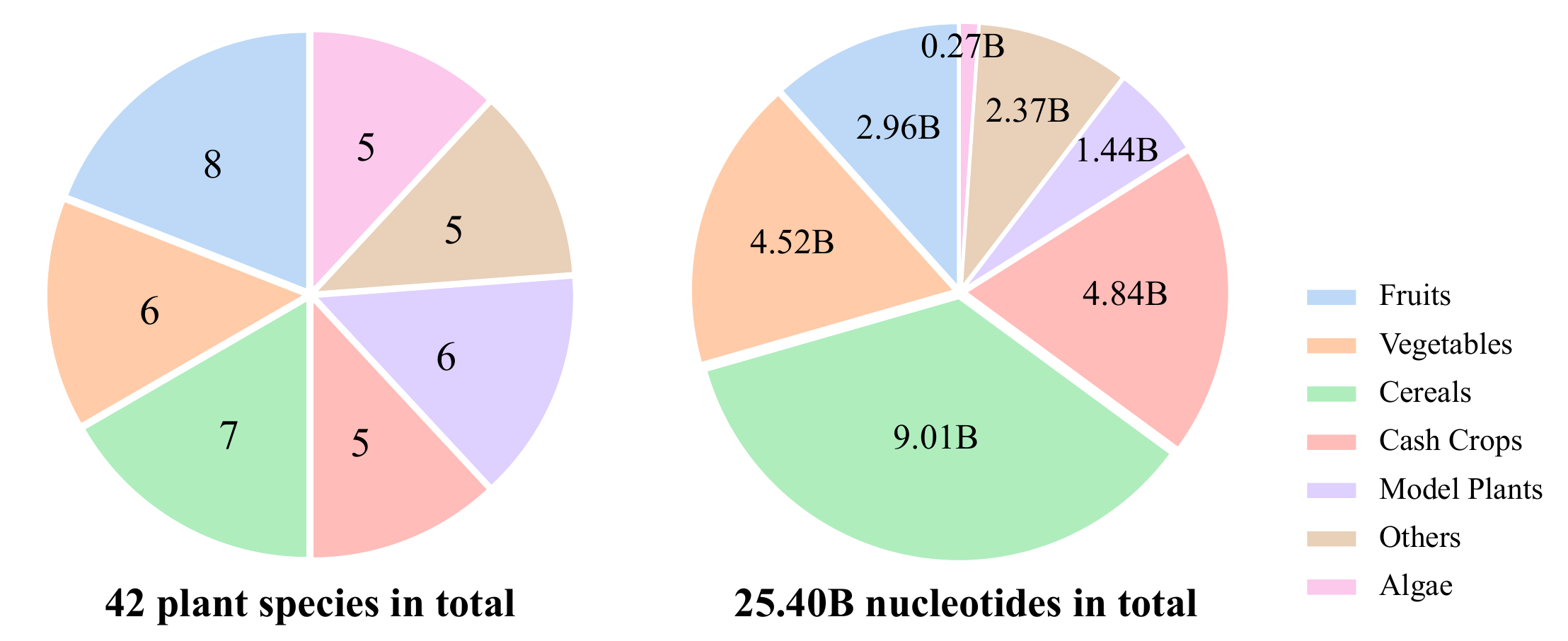}
\caption{Pre-training species categories and size for PlantBiMoE.} 
\label{fig:f0}
\end{figure}

To train PlantBiMoE, we constructed a high-quality pre-training dataset of plant genomes, which covers a total of 42 plant species in six broad categories: model plants, vegetables, fruits, cereals, algae and other important plants, as shown in Fig. \ref{fig:f0}. For each species, we obtained its reference genome FNA file from NCBI and performed the following data preprocessing sequentially: 
\begin{itemize}
    \item The reference genomes of all plants were cut into segments with a fixed length of 32,768 bp. An overlap strategy was introduced during the cut into segments to do implicit data enhancement, i.e., an overlap interval of 64 to 128 bp was retained between two neighboring segments, and the sliding step size was randomly sampled within this range.
    \item Replace all non-standard bases in the fragments, other than A, T, C, G, and N, with N uniformly.
    \item Filter out all sequences with more than 2\% of base N to form the initial dataset.
    \item Randomly select 30\% of the sequences in the initial dataset for reverse complementation (RC) and add them to the initial dataset to form the final dataset.
\end{itemize}

 Finally, we obtained a dataset containing 25.40B nucleic acid pairs. We divided the dataset into a training set and a test set according to chromosomes, with 5\% of the test set, which ensures the relative independence of the training and validation sets in terms of species and chromosomes. This strategy helps to improve the generalization assessment validity of the model in the validation stage.

2) \textit{MPGB: A Modified Plants
Genome Benchmark}

\begin{table*}[!h]
    \centering
    \caption{Summarization of the Modified Plants Genome Benchmark(MPGB).}
    \label{tab:t1}
    \makebox[0.8\textwidth][c]{%
        \begin{tabular*}{0.8\textwidth}{@{\extracolsep{\fill}}ccccc}
            \toprule
            \textbf{Task} & \textbf{No.datasets} & \textbf{No.classes / Task type} & \textbf{Sequence length} & \textbf{Source}\\
            \midrule
            Polyadenylation & 6 & 2 & 400 & \cite{AgroNT}\\
            Splicing site  & 2 & 2 & 398 & \cite{splice}\\
            LncRNA & 6 & 2 & 101--6000 & \cite{AgroNT}\\
            Enhancer region  & 1 & 2 & 1000 & \cite{AgroNT}\\
            Chromatin accessibility  & 6 & 9--19 & 1000 & \cite{plantdeepsea}\\
            Promoter strength  & 2 & Regression & 170 &  \cite{[promoter]} \\
            Terminator strength  & 2 & Regression & 170 & \cite{terminator} \\
            Histone modification  & 3 & 2 & 100--2000 & \cite{PDLLMs}\\
            Core promoter & 1 & 2 & 300 & \cite{PDLLMs}\\
            Conservation & 1 & 2 & 1000 & \cite{PDLLMs}\\
            Open chromatin & 1 & 3 & 50--2998 & \cite{PDLLMs}\\
            \bottomrule
        \end{tabular*}
    }
\end{table*}

To comprehensively assess the performance of existing plant genome language models, we integrated and constructed the Modified Plants Genome Benchmark (MPGB). The benchmark fuses AgroNT's Plants Genome Benchmark (PGB) with the tasks of Histone modification analysis, conservation identification, and core promoter identification from PDLLMs. Ultimately, the MPGB contains 11 types of tasks and 31 subdatasets covering binary, multiclassification, regression, and segmentation task types for typical model plants such as Arabidopsis thaliana, rice, maize, and soybean. Additional task details are shown in TABLE \ref{tab:t1}.

\subsection{Model Architecture}

PlantBiMoE is a 16-layer lightweight plant genome language model with an embedding dimension of 512, and its architecture is an enhancement of Jamba\cite{Jamba}, as shown in Fig. \ref{fig:f1}. We replace the original Mamba/Transformer modules with the Bidirectional Mamba (BiMamba)\cite{Caduceus} module, while retaining the interleaved SwiGLU\cite{swichglu} and SparseMoE\cite{SparseMoE} structure. Specifically, all odd layers use the BiMamba-SwiGLU module, and all even layers use the BiMamba-MoE module. This design improves parameter utilization while controlling computational overhead, ensuring scalability and practicality.

Compared to current mainstream genome language models, PlantBiMoE is the only sparse model, with a total parameter count of 116M, similar to DNABERT2\cite{DNABERT2}. However, due to its sparse structure, PlantBiMoE has only 64M active parameters per token. Furthermore, PlantBiMoE has a theoretical context window of 32,768 bp, far surpassing mainstream plant genome models such as AgroNT (6,000 bp) and PDLLMs (2,000 bp), and even exceeding DNABERT2 and NT-v2\cite{NT}, though it is smaller than HyenaDNA\cite{hyenadna} and Caduceus\cite{Caduceus}, as shown in TABLE \ref{tab:t2}. This feature enables PlantBiMoE to capture long range dependencies in the genome, which is essential for genomic language models.

\textit{1) BiMamba}

Mamba\cite{mamba} is a sequence modeling architecture based on a linear State Space Model (SSM)\cite{ssm} designed to efficiently handle long sequence tasks. Its core idea is to achieve dynamic modeling of input sequences through a selective state update mechanism. The state update and output calculation formulas of Mamba module are as follows:
\begin{equation}h_{t}=\overline{A}h_{t-1}+\overline{B}x_{t}\end{equation}
\begin{equation}y_{t}=Ch_{t}+Dx_{t}\end{equation}
where $h_{t}$ is the current hidden state and $\overline{A},\overline{B},C,D$ are the model parameters, obtained after discretization.

The BiMamba module builds on Mamba by introducing a bidirectional modeling mechanism designed to capture both forward and backward dependencies of sequences. For the input sequence $X=[x_{1},x_{2},\cdots,x_{T}] \in \mathbb{R}^{T \times d}$, Bimamba performs forward and reverse state updates, respectively:
\begin{itemize}
    \item forward:
    \begin{equation}h_{t}^{\rightarrow}=\overline{A} h_{t-1}^{\rightarrow}+\overline{B}x_{t},y_{t}^{\rightarrow}=\overline{C}h_{t}^{\rightarrow}+\overline{D}x_{t}\end{equation}
     \item reverse:
    \begin{equation}h_{t}^{\leftarrow}=\overline{A}h_{t-1}^{\leftarrow}+\overline{B}x_{t},y_{t}^{\leftarrow}=\overline{C}h_{t}^{\leftarrow}+\overline{D}x_{t}\end{equation}
\end{itemize}

After processing the forward and reverse sequences separately, this paper fuses their output representations $y_{t}^{\rightarrow}$ with $y_{t}^{\leftarrow}$. In this paper, we adopt an element-by-element addition strategy: 
\begin{equation}h_{t}=y_{t}^{\rightarrow}+y_{t}^{\leftarrow}\end{equation}
Element-wise addition is chosen due to its simplicity and computational efficiency. While more complex fusion strategies (e.g., gating or concatenation) are possible, the addition approach retains bidirectional contextual information effectively with minimal overhead.

This strategy integrates the semantic information of the two directions without introducing additional parameters, and is realized with the reversed sequences by flipping the input $ X^{\leftarrow}=flip(X,dim=1)$ to the reverse Mamba module, and the output to be flipped again to restore the temporal order.

Compared to recurrent models such as Long Short-Term Memory (LSTM) \cite{lstm} or Gate Recurrent Unit (GRU) \cite{gru}, BiMamba’s linear transition mechanism provides better gradient flow and improved memory efficiency in long-range sequence modeling.

\textit{2) SparseMoE}

Mixture-of-Experts (MoE)~\cite{moe} architectures have emerged as a scalable solution for increasing model capacity without linearly increasing computational cost. By activating only a small subset of specialized subnetworks (experts) for each input token, MoE models allow for dynamic routing of information and conditional computation, which enables efficient training of large models.

In order to enhance the expression ability and parameter utilization of the model, this paper introduces the SparseMoE\cite{SparseMoE} in the model architecture, where the module selects a number of experts for each token to be specialized through a router, so as to enhance the model modeling ability without increasing the computational cost significantly. Enhancement of model modeling capabilities without significantly increasing the computational cost. 

Specifically, given the hidden state input $\mathbf{H} \in \mathbb{R}^{B \times L \times d}$, where $B$ denotes the batch size, L is the length of the sequence, and d is the hidden dimension, it is firstly flattened into a two-dimensional matrix for expert routing: 
\begin{equation} 
\tilde {\mathbf{H}} = \mathrm{reshape}(\mathbf{H}) \in \mathbb{R}^{(B \cdot L) \times d} 
\end{equation}
Then the scores of each token for all experts are obtained by linear transformation: 
\begin{equation}
\mathbf{R} = \mathrm{softmax}(\tilde{\mathbf{H}} W_r) \in \mathbb{R}^{(B \cdot L) \times N}
\end{equation}
where $W_r \in \mathbb{R}^{d \times N}$ is the routing parameter and $N$ denotes the total number of experts. 

After that, the experts with the top $k$ scores are selected (Top-$k$ routing) and corresponding weights are assigned to each token. Top-$k$ routing not only reduces the computational burden by activating a sparse subset of experts, but also enhances specialization by allowing each expert to focus on a specific subset of the input space. This sparsity encourages diversity among experts and reduces interference between unrelated inputs, improving generalization. And each expert $E_i$ is an MLP subnetwork of the form: 
\begin{equation}
E_i(x) = W_d^{(i)} \cdot \sigma \left(W_u^{(i)} x \odot W_g^{(i)} x\right)
\end{equation}
where $\sigma(\cdot)$ denotes the SiLU\cite{silu} activation function, $\odot$ is the element-by-element multiplication, and $W_g^{(i)}, W_u^{(i)}, W_d^{(i)}$ are the gating, up-projection and down-projection parameters, respectively.

\section{Results}
\subsection{Model Training And Setup}

\textit{1) Tokenization}

PlantBiMoE’s tokenization strategy adopts the single-nucleotide method, similar to those used in HyenaDNA\cite{hyenadna} and Caduceus\cite{Caduceus}. Our base vocabulary includes the nucleotides A, T, C, G, N, and special tokens such as \textless CLS\textgreater, \textless SEP\textgreater, \textless PAD\textgreater, \textless MASK\textgreater, \textless UNK\textgreater, \textless BOS\textgreater, and \textless RESERVED\textgreater. As a result, the vocabulary size is 12.

\textit{2) Masked Lauguage Modeling}

The pre-training strategy of PlantBiMoE utilizes Masked Language Modeling (MLM)\cite{bert}. During the pretraining process, 15\% of the tokens in the input sequence are randomly masked, and the model’s task is to predict the original tokens at these masked positions. The masking mechanism follows the strategy of BERT, where 80\% of the selected tokens are replaced with the \textless MASK\textgreater, 10\% are replaced with random tokens, and the remaining 10\% are left unchanged. This strategy enhances the model's capacity to capture global contextual dependencies and, based on this, improve its ability to model sequence structures, thereby enhancing its performance across various downstream tasks.

\textit{3) Pre-training Details}

The pre-training of PlantBiMoE was distributed across a computing node with 8 Nvidia A800-80G GPUs, where the batch size for each GPU was set to 4. With 8-step gradient accumulation, the effective batch size became 256. The AdamW optimizer\cite{adamw} was used, with $\beta_{1}$ set to 0.95, $\beta_{2}$ to 0.9, and a weight decay of 0.1. The total number of training steps was equivalent to 10 epochs. During the initial 2\% of the training steps, the learning rate increased linearly from 0 to 0.008, followed by a cosine decay to 0.004. Mixed precision training with \texttt{bfloat16} (bf16) was adopted to improve training efficiency and reduce memory overhead, resulting in a total pre-training time of approximately 166 hours.

\subsection{Baseline}

For tasks previously reported by AgroNT and PDLLMs, we directly utilize the results published in their respective papers. For tasks not covered by AgroNT, we perform parameter-efficient fine-tuning using the IA3\cite{IA3} method, adhering to the methodology described in their work. For tasks unreported by PDLLMs, we employ the PlantDNAMamba series of models—including variants such as PlantDNAMamba-singlebase, PlantDNAMamba-6mer, and PlantDNAMamba-BPE under full-parameter fine-tuning. This selection is made because the PlantDNAMamba series demonstrated the highest average performance across PDLLMs’s evaluations\cite{PDLLMs} (results from these models are collectively denoted as ``PlantDNAMamba'' throughout this paper). For all tasks, we exclusively use PlantDNAMamba-6mer for fine-tuning. This decision is based on our evaluation, which indicated that PlantDNAMamba-singlebase and PlantDNAMamba-BPE underperformed in certain classification and regression tasks.

\begin{table}[!t]
    \centering
    \caption{Comparison of Model params and context length}
    \label{tab:t2}
    \begin{tabular}{cccc}
        \toprule
        \textbf{Model} & \boldmath{$N_{params}$} & \boldmath{$N_{act-params}$} & \textbf{context length}\\
        \midrule
        NT-v2-multi & 500M & 500M & 12,000 \\
        DNABERT-2 & 117M & 117M & 10,000 \\
        HyenaDNA & 7M & 7M & 160,000 \\
        Caduceus &  8M & 8M &  131,072\\
        AgroNT-1B & 1B & 1B & 6,000 \\
        PlantDNAMamba & 94M & 94M & 2,000\\
        PlantBiMoE  & 116M & 64M & 32,768 \\
        \bottomrule
    \end{tabular}
\end{table}

\subsection{Metrics}

For different task types, we use the appropriate evaluation metric. The $R^{2}$ coefficient is used uniformly for all regression tasks. For the classification tasks (including Polyadenylation, Splice sites, LncRNA, Chromatin accessibility and Enhancer region) in the PGB benchmark constructed by AgroNT, we follow the Area Under Curve (AUC) reported in the original paper as the evaluation metric. For the classification tasks in the PDLLMs benchmark (including Histone modification, Core promoter, Conservation, and Open chromatin), we chose MCC instead of the F1 score used in the original paper. This adjustment is due to the fact that MCC provides a more comprehensive assessment of classification performance and the results are provided in the PDLLMs.

\begin{table*}[!t]
    \centering
    \caption{The models’ averaged performance on the 11 tasks in the MPGB}
    \label{tab:t3}
    \begin{threeparttable}
    \begin{tabular}{ccccc}
        \toprule
        \textbf{Task} & \textbf{Metrics} & \textbf{AgroNT /} \boldmath{$N_{top1}$}\tnote{1} & \textbf{PlantDNAMamba /}  \boldmath{$N_{top1}$}\tnote{1} & \textbf{PlantBiMoE /}  \boldmath{$N_{top1}$}\tnote{1}  \\
        \midrule
        Polyadenylation & AUC & 93.62 / 1 & 90.22 / 4 & \textbf{93.63} / 1 \\
        Splicing site & AUC & 99.81 / 0 & 99.79 / 0 & \textbf{99.84} / 2 \\
        LncRNA & AUC & 83.11 / 1 & 82.79 / 1 & \textbf{84.12} / 4 \\
        Enhancer region & AUC & \textbf{88.15} / 1 & 84.04 / 0 & 87.47 / 0 \\
        Chromatin accessibility & AUC & 96.37 / 1 & 96.54 / 1 & \textbf{96.55} / 4 \\
        Promoter strength & $R^{2}$ & 73.85 / 0 & 73.35 / 0 & \textbf{75.23} / 2 \\
        Terminator strength & $R^{2}$ & 71.66 / 0 & 69.38 / 0 & \textbf{72.22} / 2 \\
        Histone modification & MCC & 63.99 / 0 & 66.01 / 0 & \textbf{66.27} / 3 \\
        Core promoter & MCC & 58.74 / 0& \textbf{64.41} / 1 & 63.20 / 0\\
        Conservation & MCC & 80.14 / 0 & 81.18 / 0 & \textbf{81.74} / 1 \\
        Open chromatin & MCC & 43.21 / 0 & 46.24 / 0 & \textbf{46.88} / 1 \\
        \midrule
        Total Top & / &  4 &  7 & \textbf{20} \\
        \bottomrule
    \end{tabular}
    \begin{tablenotes}    
        \footnotesize               
        \item[1] Left: Average score per model; Right: Count of top-1 rankings across subtasks.
      \end{tablenotes}  
    \end{threeparttable} 
\end{table*}

\subsection{Results on MPGB}
The average performance of all models on the MPGB benchmarks is summarized in TABLE \ref{tab:t3}. PlantBiMoE achieves the best average performance on 9 out of 11 tasks and on 20 out of 31 subdatasets across all tasks. Notably, despite the fact that PlantBiMoE's model size is 10 times smaller than AgroNT's and does not employ a PlantDNAMamba-like multi-model selection strategy, its overall performance on MPGB (except for the tasks of Enhancer region, and Core promoter) are all better than AgroNT and PlantDNAMamba, fully reflecting the model's advantages in performance and efficiency. 

\begin{figure}[bthp]
\centering
\includegraphics[width=0.48\textwidth]{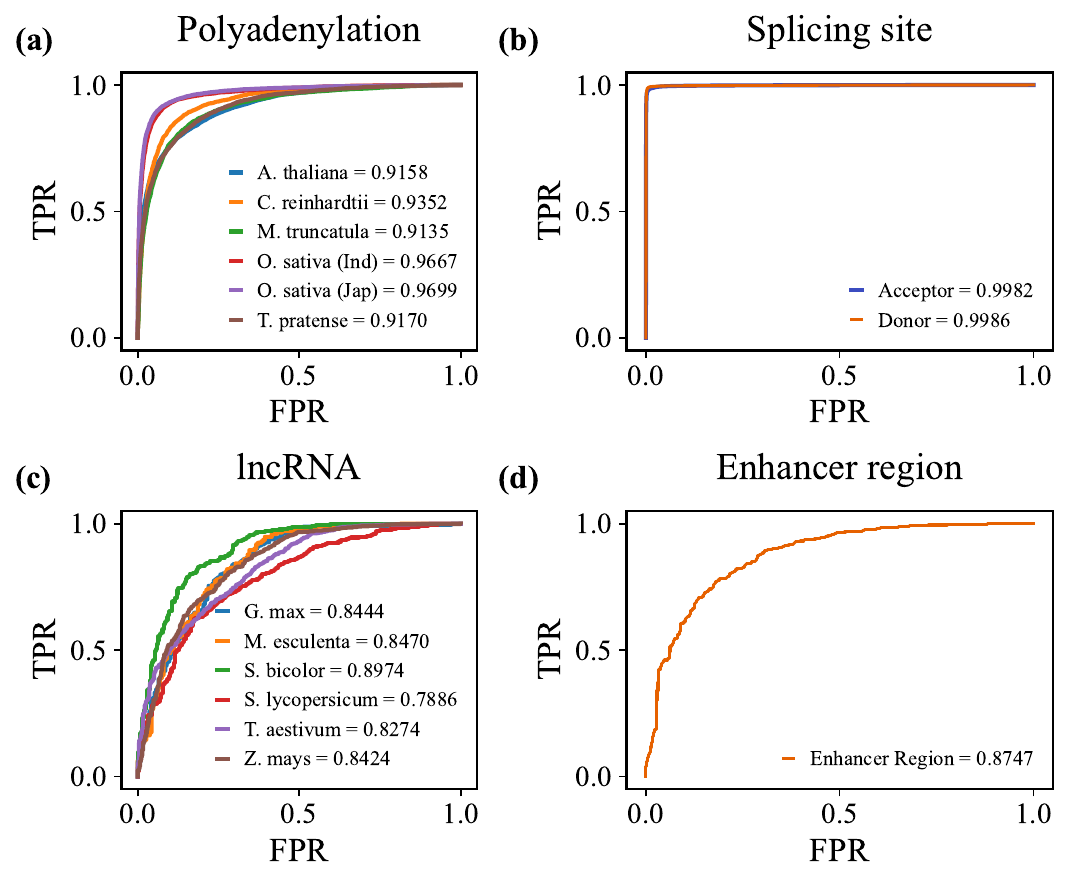}
\caption{PlantBiMoE predicts Polyadenylation, Splicing site, LncRNA and Enhancer region. (a)Receiver operating characteristic curve for Polyadenylation prediction . (b) Same as in a, but for Splicing site prediction. (c) Same as in a, but for LncRNA. (d) Same as in a, but for Enhancer region. } 
\label{fig:f2}
\end{figure}

\begin{figure}[bthp]
\centering
\includegraphics[width=0.50\textwidth]{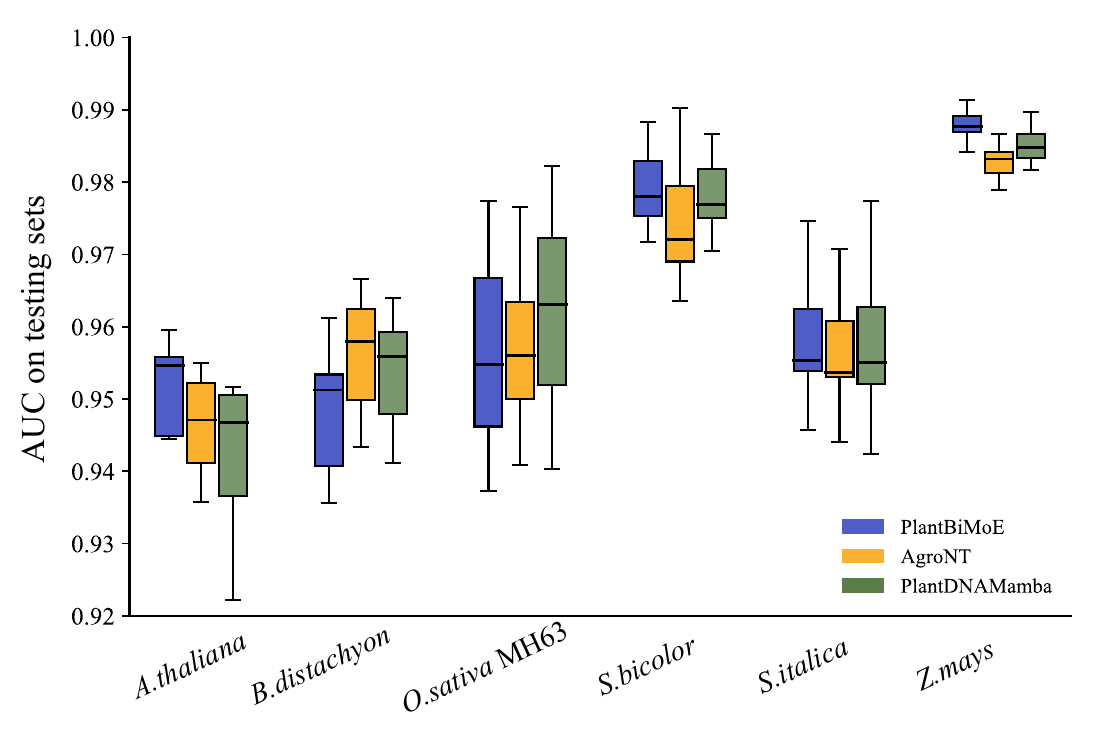}
\caption{Box plot of Chromatin accessibility task for six species.} 
\label{fig:f3}
\end{figure}

As can be seen from the Fig. \ref{fig:f2}, the AUC curve has increased significantly when the false positive rate (FPR) is low, which indicates that the model has a strong early identification ability and can effectively distinguish between positive and negative samples with fewer misjudgments. Boxplots of AUC distributions across six plant species were plotted to compare the chromatin accessibility prediction performance of PlantBiMoE, AgroNT, and PlantDNAMamba, as shown in Fig. \ref{fig:f3}. The results show that PlantBiMoE achieves competitive or superior performance on most species, with particularly high AUC scores and reduced variance on \textit{Z.mays}, \textit{S.bicolor} and \textit{A.thaliana}. These results suggest that PlantBiMoE offers improved generalization and robustness in cross-species settings, although performance advantages may vary depending on the species and specific testing sets.

Specifically, in the six tasks of Splicing sites, Promoter strength, Terminator strength, Histone modification, Conservation, and Open chromatin, PlantBiMoE perform better than AgroNT and PlantDNAMamba across all 11 subdatasets. On both the LncRNA and Chromatin accessibility tasks, PlantBiMoE achieve the best performance on 4 of the 6 subdatasets and the best average task performance (84.12 and 96.55), respectively. On the Polyadenylation task, while PlantDNAMamba perform best on 4 of the 6 sub-datasets, its result on the \textit{M.truncatula} subdataset (69.60) is significantly lower than those of AgroNT (94.57) and PlantBiMoE (91.35), which resulting in a lower average performance (90.22) than AgroNT (93.62) and PlantBiMoE (93.63) on this task as well. On the Enhancer region and Core promoter tasks, PlantBiMoE's performance is slightly lower than the best model, but still stabilize at the second place among the three models, as shown in TABLE \ref{tab:t3}.

\section{Discussion and Conclusion}

In this study, we introduce PlantBiMoE, a novel lightweight language model tailored for plant genomes. PlantBiMoE combines BiMamba modeling with SparseMoE framework to maintain strong representation capabilities while significantly enhancing computational efficiency. It demonstrates excellent performance and cross-species generalization across 11 representative downstream tasks, including genome structure prediction, transcriptional regulation, and epigenetic modeling. Compared with existing models such as AgroNT and PDLLMs, PlantBiMoE consistently achieves superior results, establishing a new benchmark for lightweight plant genome models.

These results strongly suggest that the effectiveness of genomic language models lies not merely in scaling up model size or pre-training data volume, but in the careful integration of architectural design and data quality. This insight offers practical guidance for the development of efficient, scalable models for genome analysis in the future.

The main contributions of this work are as follows: (1) We are the first to integrate BiMamba and SparseMoE into a plant genome language model, achieving strong sequence representation capability. (2) The resulting pre-trained genome language model, built upon these two key innovations, lays a transformative foundation for various aspects of plant genome research. (3) We propose the Modified Plants Genome Benchmark (MPGB), a unified and enhanced benchmark comprising 31 datasets spanning diverse genomic tasks with sequence lengths ranging from 50 to 6,000 bp.

\section*{funding}

This research was supported by the National Key Research and Development Program of China (2023YFD1202903).

\bibliographystyle{ieeetr}
\bibliography{ref}

\end{document}